%% file: main.tex
\documentclass[10pt,twocolumn,letterpaper]{article}

\usepackage{cvpr} 


\usepackage{siunitx}

\usepackage{epsfig}
\usepackage{amsmath}
\usepackage{amssymb}
\usepackage{booktabs}
\usepackage{multirow}
\usepackage{multicol}
\usepackage[ruled,vlined]{algorithm2e}
\usepackage{pifont}
\usepackage[dvipsnames, svgnames, x11names]{xcolor}         

\usepackage{colortbl}       
\usepackage{pifont}
\usepackage{graphicx} 

\usepackage{algorithmic}
\usepackage[dvipsnames, svgnames, x11names]{xcolor}         
\newcommand{\website}{\url{https://activeumi.github.io}}

\definecolor{cvprblue}{rgb}{0.21,0.49,0.74}
\usepackage[pagebackref,breaklinks,colorlinks,allcolors=cvprblue]{hyperref}

\def\paperID{4051} 
\def\confName{CVPR}
\def\confYear{2025}

\title{ActiveUMI: Robotic Manipulation with Active Perception \\ from Robot‑Free Human Demonstrations}

\author{
\text{Qiyuan Zeng$^{1,}$}\thanks{: Co-first author. $\dagger$: Corresponding Author. This work was done during Qiyuan Zeng’s internship at Midea Group.} \text{, Chengmeng Li$^{1}$}, 
\text{Jude St. John$^{2,\dagger}$}, 
\text{Zhongyi Zhou$^{3}$}, \\ \text{Junjie Wen$^{3}$}, \text{Guorui Feng$^{1,\dagger}$}, \text{Yichen Zhu$^{3,*,\dagger}$}, \text{Yi Xu$^{3}$}
\\
\tt\small $\textsuperscript{1}\text{Shanghai University}$, $\textsuperscript{2}\text{Stanford University}$, $\textsuperscript{3}\text{Midea Group}$
\\
\hspace{0cm}\large\website
\vspace{-0.5cm}
}

\begin{document}

\makeatletter
\let\@oldmaketitle\@maketitle
\renewcommand{\@maketitle}{\@oldmaketitle
    \begin{center}
        \captionsetup{type=figure}
        \centering
        \includegraphics[width=0.9\textwidth]{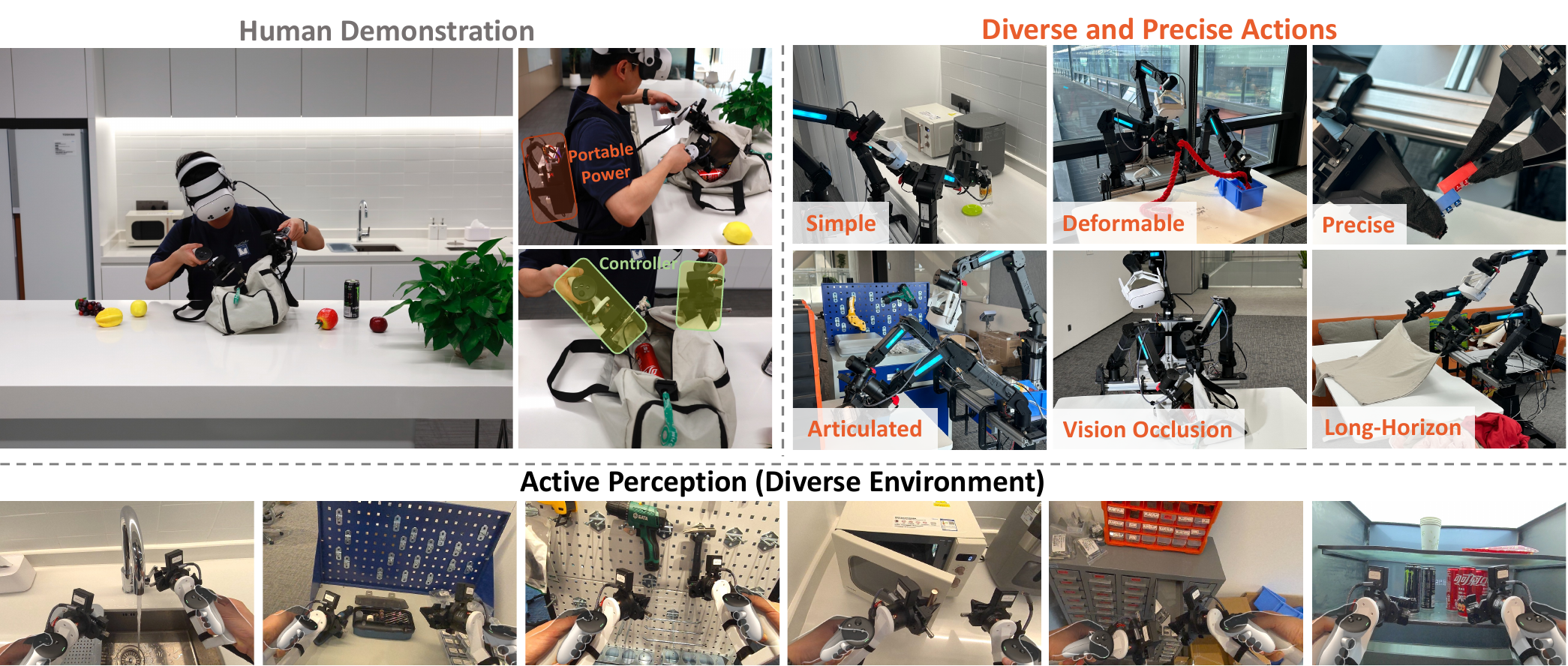}
        \caption{\textbf{Universal Manipulation Interface with Active Perception (ActiveUMI)} is a portable, low-cost data collection framework for transferring diverse, in-the-wild human demonstrations into effective visuomotor policies. The core of our method is to empower this system with active perception, which allows the robot to control its viewpoint. This capability is critical for completing long-horizon tasks, overcoming visual occlusions, and performing actions that require high precision.}\label{fig:system}
    \end{center}
}
\makeatother

\maketitle

\begin{abstract}
We present ActiveUMI, a framework for a data collection system that transfers in-the-wild human demonstrations to robots capable of complex bimanual manipulation. ActiveUMI couples a portable VR teleoperation kit with sensorized controllers that mirror the robot's end-effectors, bridging human-robot kinematics via precise pose alignment. To ensure mobility and data quality, we introduce several key techniques, including immersive 3D model rendering, a self-contained wearable computer, and efficient calibration methods. ActiveUMI's defining feature is its capture of active, egocentric perception. By recording an operator's deliberate head movements via a head-mounted display, our system learns the crucial link between visual attention and manipulation. We evaluate ActiveUMI on six challenging bimanual tasks. Policies trained exclusively on ActiveUMI data achieve an average success rate of 70\% on in-distribution tasks and demonstrate strong generalization, retaining a 56\% success rate when tested on novel objects and in new environments. Our results demonstrate that portable data collection systems, when coupled with learned active perception, provide an effective and scalable pathway toward creating generalizable and highly capable real-world robot policies.
\end{abstract}

\section{Introduction}
Robot foundation models promise generalist policies but are currently constrained by the scale and alignment of available robot data relative to web‑scale corpora. A central challenge is therefore scaling data collection while preserving embodiment fidelity. Prevailing sources—in‑lab teleoperation, human videos, and simulation—each have limitations: teleoperation is costly to scale; human videos~\cite{bahl2022human, qiu2025humanoidpolicy, fu2024humanplus, ye2025video2policy, hu2024vpp} incur a cross‑embodiment gap (human to robot); and simulation suffers a sim‑to‑real gap (physics to hardware~\cite{maddukuri2025simandreal}).

A promising middle ground is sensorized hand‑held interfaces (e.g., grippers, dexterous‑hand devices) that capture action‑aligned trajectories. Yet most current interfaces overlook active, egocentric perception: humans move their heads to manage occlusion and gather context, while existing rigs rely primarily on wrist‑mounted cameras. Even with a wide field-of-view, an end‑effector–centric view underserves long‑horizon tasks and fine manipulation and misaligns with platforms that use head‑mounted cameras. These observations motivate data‑collection and policy‑learning pipelines that couple head‑ego sensing with wrist‑eye control, enabling viewpoint selection as part of the task and improving transfer to real robots.

To this end, we propose ActiveUMI, a universal manipulation interface with active perception for in-the-wild robot policy learning. Our approach is built on two core principles for scalable data collection: (i) the system must tightly align the robot's embodiment with natural human movement, and (ii) it must enable active perception to expose the right sensory information at the right time. Our system addresses these needs with a specially designed, portable VR teleoperation kit. We developed a hardware architecture that allows the target robot's own custom grippers to be mounted directly onto the VR controllers, mirroring the end-effectors precisely. The entire system is self-contained in a backpack, and we implement several calibration techniques to ensure consistent, high-quality data collection in diverse real-world environments. To enable active perception, we map the operator's head movements to a movable robotics arm with a head-mounted camera. This allows the learned policy to control its own viewpoint, actively seeking out information to solve complex, long-horizon, or visually occluded tasks that are challenging for systems with only static or wrist-mounted cameras.

We evaluate ActiveUMI on six challenging, real‑robot bimanual tasks that combine precise hand–object interactions with long‑horizon manipulation using only the egocentric head camera and wrist proprioception available to the robot platform. By training policies trained purely on ActiveUMI demonstrations, they attain an average 70\% success rate on all tasks. Relative to non-active perception counterparts (i.e., policies trained from wrist‑centric views or static third‑person cameras), ActiveUMI improves average success by 44\% and 38\%, respectively. Furthermore, when evaluated with novel objects and scenes, learned policies retain 56\% of the average success rate, indicating a meaningful generalization from in-wild data.
\begin{figure*}[t]
    \centering
    \includegraphics[width=0.95\linewidth]{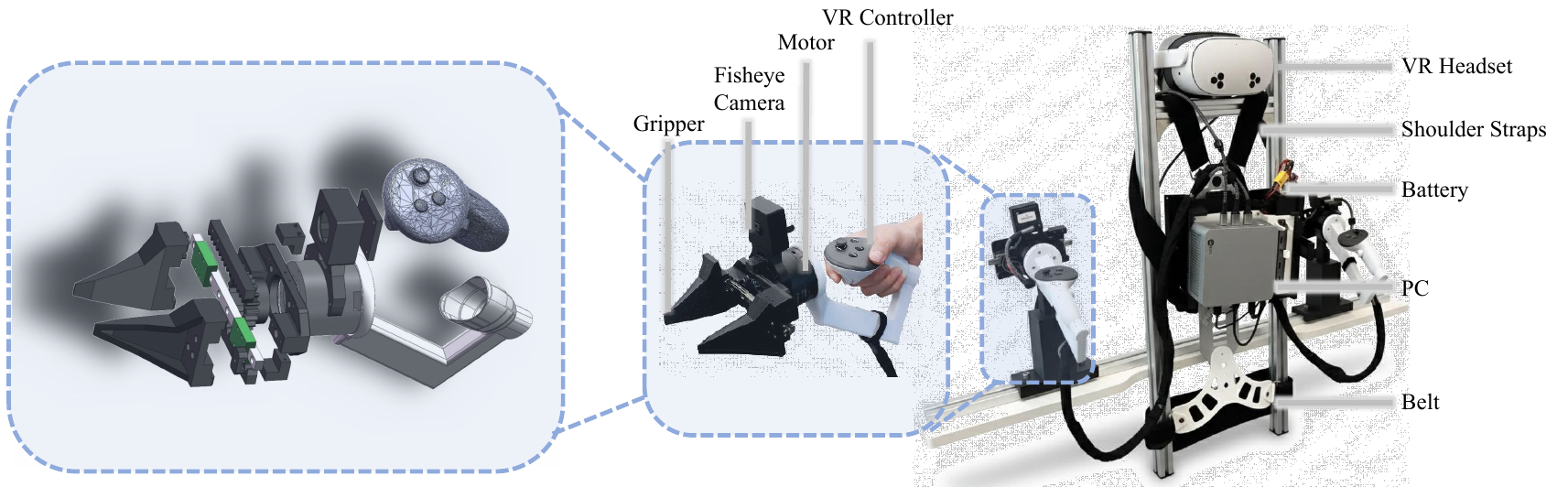}
    \caption{\textbf{Overview of ActiveUMI Hardware.} A VR headset with custom controllers designed to replicate the structure of the robot's grippers. A portable backpack that holds a battery and a PC for self-contained operation.}
    \vspace{-1em}
    \label{fig:system_architecture}
\end{figure*}

\section{Related Work}
Data collection is a central pillar of modern deep learning, especially in the era of large models with massive numbers of learnable parameters. In robotics, the development of robot foundation models~\cite{act, chi2023diffusion_policy, ze20243d,he2024omnih2o,jiang2025dexmimicgen,jia2024lift3d,chen2025acdit,jia2024lift3d,reuss2024multimodal,dasari2024ditpi}, such as Vision-Language-Action (VLA) models~\cite{black2024pi0, intelligence2025pi05, nvidia2025gr00tn1, ding2025humanoidvla,wen2024tinyvla,zhou2025chatvla,zhou2025chatvla2,team2025gemini,li2025controlvla,deng2025graspvla,li2025pointvla,chen2025fast,liu2025hybridvla,wang2025trackvla}, has recently garnered significant attention. A critical prerequisite for training a robust and useful robot foundation model is the collection of massive datasets. However, the scale of today's robotics data is only a small fraction of that used for training large language models. Several approaches aim to alleviate this data scarcity problem, including designing user-friendly teleoperation systems~\cite{ze2025twist,sun2025ulc,li2025clone,ben2024homie,cheng2024opentelevision,fu2024humanplus}, leveraging large-scale simulation data~\cite{maddukuri2025simandreal,arachchige2025sail,geng2025roboverse}, and repurposing human videos~\cite{grauman2022ego4d, Kareer2025EgoMimic,yang2025egovla, hu2024vpp, ye2025video2policy, liu2025immimic, zhu2025emma, punamiya2025egobridge,kumar2024hrp}. However, each has significant drawbacks: teleoperation is expensive and difficult to scale, while both simulation and human videos suffer from significant reality and embodiment gaps, respectively.

To overcome the scaling limitations of in-lab setups, research has explored collecting data ``in-the-wild". One common source is using human demonstrations. DexCap~\cite{wang2024dexcap} uses a wearable glove to capture precise wrist and fingertip poses for dexterous tasks. AirExo\cite{fang2024airexo,fang2025airexo2} leverages low-cost hardware with direct kinematic mapping for arm manipulation. DoGlove~\cite{zhang2025doglove} uses a low-cost, precise, and haptic force feedback glove system for teleoperation and manipulation. Dexop~\cite{fang2025dexop} uses a passive hand exoskeleton designed to maximize human ability to collect rich sensory data for diverse dexterous manipulation tasks in natural environments. NuEXO~\cite{zhong2025nuexo} designs a portable exoskeleton hardware to do both teleoperation and collect humanoid data. The Universal Manipulation Interface (UMI)~\cite{chi2024umi} is the most related work to us. The UMI introduced a simple handheld controller for collecting bimanual data at scale, which DexUMI~\cite{xu2025dexumi} later extended to dexterous hands with similar concepts. FastUMI~\cite{liu2024fastumi} uses a substantial redesign of the UMI system that addresses these challenges by enabling rapid deployment via adding an extra camera on top of the UMI gripper. However, a common limitation among these systems is their primary reliance on wrist-mounted cameras for perception. Because these cameras move with the arm, their viewpoints are constrained by manipulation needs rather than by perceptual objectives. Vision-in-Action~\cite{xiong2025vision} is a closely related work that focuses on designing a teleoperation system for active perception. Our core contribution is the integration of active, egocentric perception by explicitly tracking the operator's head movements via a VR headset. This allows the learned policy to actively control its own viewpoint—a capability that is critical for overcoming occlusions and successfully completing complex tasks.

\section{Method}
This section introduces ActiveUMI, a high-mobility framework designed for large-scale, in-the-wild robot learning. We will first provide an overview of the data collection system, then delve into the core concept of active perception, and conclude with the calibration methods that ensure high-quality data.
\begin{figure*}[t]
    \centering
    \includegraphics[width=\linewidth]{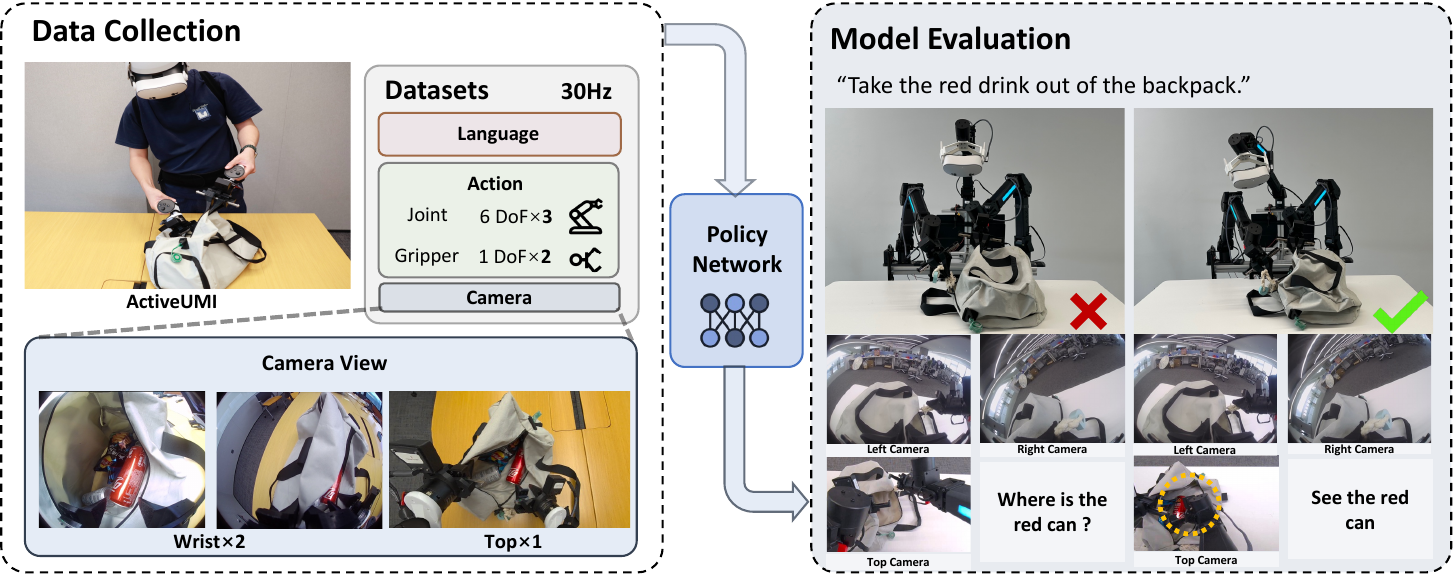}
    \caption{\textbf{Overview of ActiveUMI.} The left side of the figure illustrates our data collection process and the detailed dataset configuration. The training data from in-the-wild data collected by ActiveUMI. The right side of the figure shows the model deployment and inference process.}
    \vspace{-1em}
    \label{fig:flow_chart}
\end{figure*}

\subsection{Data Collection System for ActiveUMI}
Data collection is a central pillar of modern deep learning, especially in the era of large models with massive numbers of learnable parameters. In robotics, the development of robot foundation models~\cite{act, chi2023diffusion_policy, ze20243d}, such as Vision-Language-Action (VLA) models~\cite{black2024pi0, wen2025dexvla}, has recently garnered significant attention. A critical prerequisite for training a robust and useful robot foundation model is the collection of massive datasets. However, the scale of today's robotics data is only a small fraction of that used for training large language models. Several approaches aim to alleviate this data scarcity problem, including designing user-friendly teleoperation systems, leveraging large-scale simulation data~, and repurposing human videos. However, each has significant drawbacks: teleoperation is expensive and difficult to scale, while both simulation and human videos suffer from significant reality and embodiment gaps, respectively.

The design of ActiveUMI facilitates an intuitive and efficient process for high-quality data collection while extending the operational boundaries from constrained laboratory settings to diverse, ``in-the-wild" environments. To this end, we have developed a low-cost, high-precision hardware system based on consumer-grade VR equipment, with its overall architecture depicted in Figure~\ref{fig:system_architecture}.

\textbf{VR gripper controller.} Our VR controller is a modified version of the commercial Meta Quest 3s controller, leveraged for its inherent capability for synchronous, low-latency, and high-precision six-degrees-of-freedom (6-DoF) pose tracking. This is accomplished via the headset's sophisticated inside-out tracking system. The headset's onboard cameras continuously triangulate the controller's pose in real-time by tracking a unique pattern of integrated infrared (IR) LEDs. By obtaining the 6-DoF pose data, we can concurrently resolve both the controller's translational position (x,y,z) and its rotational orientation (roll, pitch, yaw) within the captured volume. Consequently, by rigidly mounting this controller onto our target robot, its pose becomes directly representative of the robot's pose. A detailed analysis of the measurement error is provided in Section~\ref{sec:throughput_and_accuracy}. 

Our approach offers greater hardware flexibility compared to systems like UMI, which are often built around a specific, non-interchangeable gripper. We can adapt our system by simply mounting a modified Meta Quest controller onto the target robot's existing end-effector.

\textbf{Gripper actuation.} We integrate a micro-motor directly onto the controller to drive the open-close motion of the gripper. This allows an operator to control the robot's grasp intuitively. A key advantage of our design is that it's non-invasive; instead of replacing the robot's ``vanilla" gripper, we attach an identical copy to the operator's controller for data collection. This ensures our system can be deployed on a wide range of stock robots with minimal modification.

To enrich the data stream, we augment each controller with a fisheye camera.  This wrist-mounted camera is positioned to maximize its field of view, capturing comprehensive visual information of the robot's immediate operational environment. This provides the downstream policy model with rich visual context, and the resulting ``wrist view" serves as a valuable complement to the first-person perspective from the head-mounted camera.

\textbf{Head-mounted display (HMD).} The Meta Quest3s HMD plays a dual, critical role within our framework. Firstly, it serves as a high-precision localization hub. Its robust SLAM system provides a stable and reliable world coordinate system, concurrently tracking the 6-DoF poses of both the operator's head and the controller. Secondly, the HMD's front-facing color cameras function as a dynamic, top camera, offering a global perspective that is intrinsically coupled with the operator's line of sight.

\textbf{Wearable device.} To enable data collection in any environment, we utilize a compact, wearable computational unit consisting of a small computer worn on the operator's back. This self-contained design liberates the operator from a stationary workstation, allowing them to move freely and gather data across diverse settings.

\begin{figure}[t]
\centerline{\includegraphics[width=0.9\linewidth]{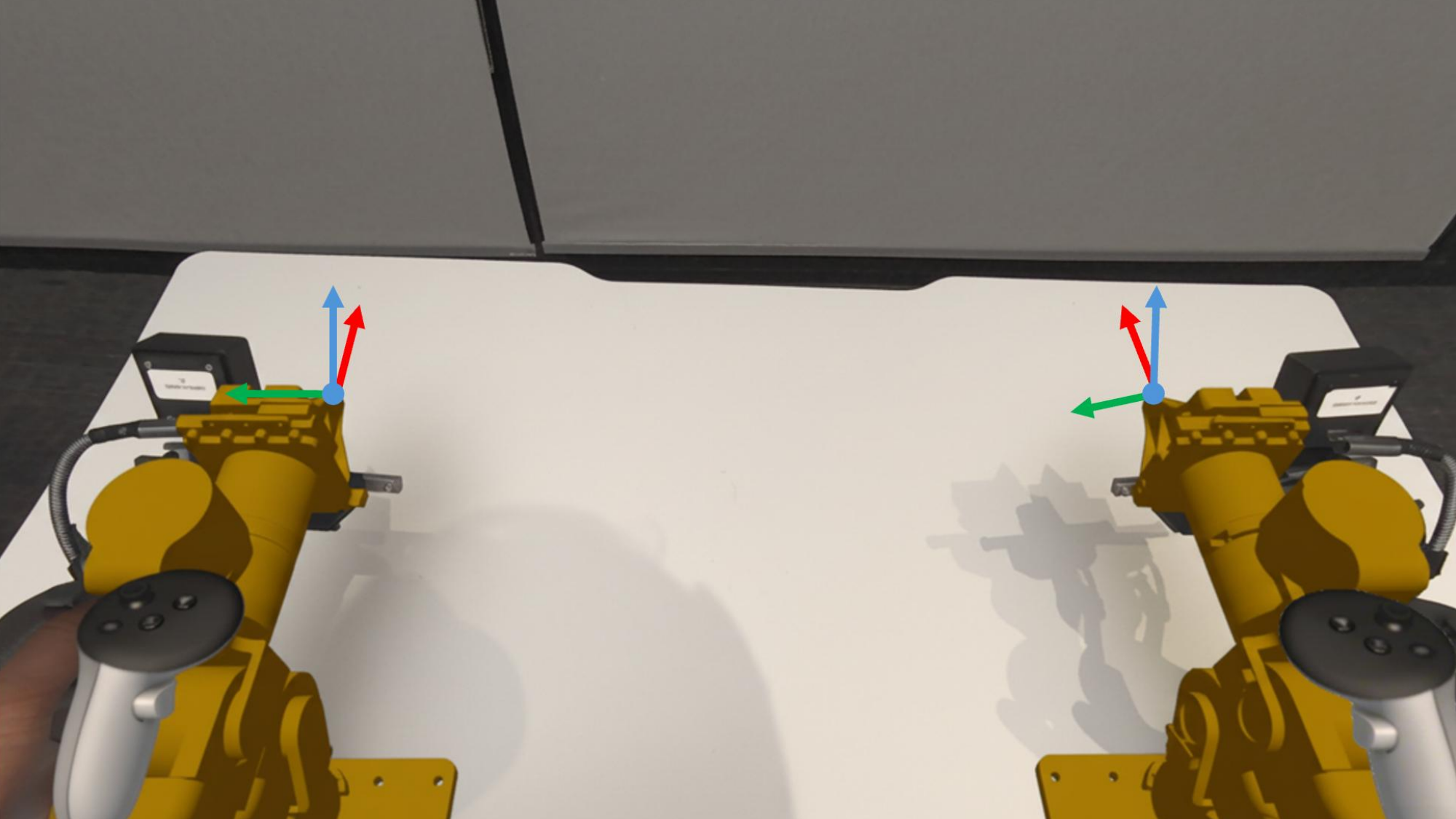}}
\caption{\textbf{Immerse Data Collection.} Our system provides the operator with critical visual feedback by rendering the robot's arms in the VR environment.}
\label{fig:immerse_data_collect}
\end{figure}

\textbf{Immerse data collection.} To provide the operator with intuitive feedback, we render a 3D model of the robotic arms within the VR environment. These virtual arms are precisely aligned with the operator's hand-held controllers, which correspond to the robot's grippers. This setup allows the operator to clearly visualize the robot's movements in real-time during data collection. We visualize the rendered model in Figure~\ref{fig:immerse_data_collect}.

\begin{figure*}[t]
    \centering
    \includegraphics[width=\linewidth]{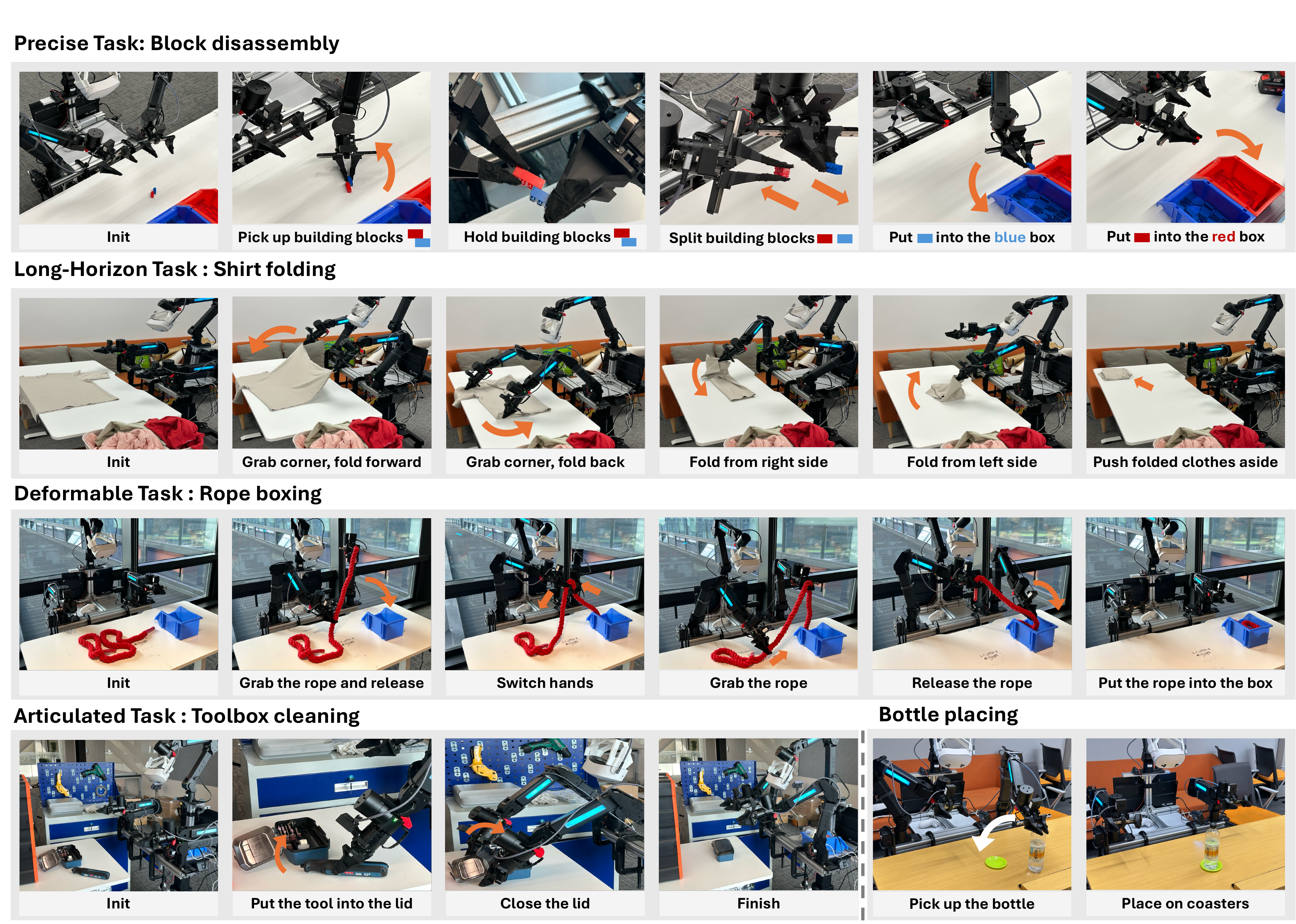}
    \caption{\textbf{Evaluated Tasks.} We evaluated our approach on a diverse set of tasks, each requiring a different skill set: \textbf{Block disassembly} is a precision task where the robot must separate two small, interlocked blocks and then sort them into a box. \textbf{Shirt folding} is a deformable object manipulation task that demands accurate state recognition to correctly fold the cloth. \textbf{Rope boxing} is a long-horizon task where the robot must neatly place a long rope into a box. \textbf{Toolbox cleaning} is an articulated object manipulation task that requires the robot to close the lid. \textbf{Bottle placing} is a task designed to test the policy's robustness to large positional variations of the objects.}
    \label{fig:tasks}
\end{figure*}

\subsection{Active Perception for Policy Learning}
A key limitation of conventional UMI-style data collection is its reliance on wrist-mounted cameras. Because these cameras move with the robot's arms, their viewpoints are constrained by manipulation needs rather than guided by perceptual objectives. This makes it difficult for a trained policy to handle scenarios with visual occlusions, manipulate deformable objects, or perform tasks that require significant shifts in viewpoint.

ActiveUMI is designed to bridge this visual gap by enabling the robot to act with human-like flexibility in its head and camera control. To achieve this, we explicitly record the real-time 6-DoF pose of the operator's Head-Mounted Display (HMD) as an additional input to the policy. This allows the model to learn the crucial correlation between an operator's head movements (i.e., their visual attention) and their corresponding hand actions.

During deployment, the policy can then predict a 6-DoF pose for the robot's head, allowing it to actively mimic the operator's learned attention patterns. This predicted motion is executed by the robot’s low-level controller, enabling the robot to dynamically adjust its viewpoint, overcome occlusions, and significantly enhance its performance on complex tasks.

\subsection{Calibrating End-Effector for Precise Data Collection}

The ActiveUMI system captures 6-DoF (Degrees of Freedom) pose data from three key points in the VR setup: the tips of the left and right controllers and the pose of the Quest 3 headset. During policy execution, these tracked points map one-to-one with the robot's two gripper tips and its head-mounted camera. All data is recorded in absolute coordinates relative to a unified world coordinate system that is established during an initial calibration phase. This ensures the reference frame remains consistent throughout the data collection session.

To ensure high-quality data alignment and maintain precision, we introduce three additional approaches to facilitate robust calibration.

\textbf{In-Situ environment setup.} To reset the 6-DoF zero-point, operators can press the `B' button on the controller to reposition the base coordinate system. This feature enables data collection to start flexibly in any environment. The coordinate system's axes are rendered in real-time within the headset, allowing the operator to intuitively align the virtual reference frame with the physical workspace. This process ensures a consistent initial state for every data collection session.

\textbf{Gripper placeholder.} To simplify calibration, we designed a physical placeholder that serves as a docking station for the VR controllers. This jig can be placed anywhere in the workspace to establish a consistent starting point. When the controllers are seated in the placeholder, their relative distance and pose are fixed to a predefined state. Pressing a designated button while the controllers are docked instantly calibrates the virtual coordinate system, aligning its origin and orientation with this known physical configuration.

\textbf{Haptic feedback for zero-point position.} To enhance the efficiency and convenience of zero-point calibration, we implemented a haptic feedback mechanism. Specifically, when a gripper moves within 3cm of the zero-point (the origin of the base coordinate system), the controller's motor generates a high-frequency vibration. This tactile cue alerts the operator that the gripper is approaching its base position. This mechanism allows users to confirm alignment without relying on numerical readouts, significantly improving the speed and efficiency of the calibration process.

By implementing the methods described above, we ensure that every data collection session begins from a precise and consistent initial pose. This guarantees an accurate one-to-one mapping between the operator's controls and the real robot's kinematics from the very start. Furthermore, these streamlined calibration procedures significantly reduce the operator's cognitive load. Ultimately, this user-centric design makes the data collection process more efficient and leads to higher-quality, more natural demonstrations, which is crucial for building a scalable framework for effective policy learning.

\section{Experiment}
In this section, we will discuss the effectiveness of our proposed ActiveUMI. Specifically, we aim to investigate the following question:
\begin{itemize}
    \item How important is the egocentric active perception for in-the-wild robot learning?
    \item What is the optimal strategy for utilizing ActiveUMI data to maximize end-to-end model performance?
    \item Can ActiveUMI data help the model generalize to new objects and scenes?
\end{itemize}

\begin{table*}[h]
\centering
\caption{We compare our active perception approach to two variants: a fixed top-down camera and a wrist-camera-only setup. The wrist-camera-only configuration corresponds to the UMI setting.}
\label{tbl:main_indomain}
\resizebox{0.95\textwidth}{!}{\begin{tabular}{l|c|c|c|c|c|c}
\toprule
\multirow{2}{*}{Camera View} & \multicolumn{6}{c}{Tasks (\textbf{In-Domain})}  \\
& Bottle placing & Rope boxing  & Shirt folding & Block disassembly & Take Drink from Bag& Average \\
\midrule
UMI & 60\%  & 20\%  & 10\%  & 0\%   & 40\%  & 26\%\\
UMI w/ Fixed Head Camera & 60\% & 40\%  & 40\% & 20\%  & 50\%  & 42\% \\
\textbf{ActiveUMI}      & \textbf{90\%}  & \textbf{70\%}  & \textbf{80\%}  & \textbf{30\%}   & \textbf{80\%} & \textbf{70\%}\\
\bottomrule
\end{tabular}}
\end{table*}

\begin{table*}[ht]
\centering
\caption{We compare our active perception approach to two variants in a new environment under the same task as Table~\ref{tbl:main_indomain}.}
\label{tbl:main_generalization}
\resizebox{0.95\textwidth}{!}{\begin{tabular}{l|c|c|c|c|c|c}
\toprule
\multirow{2}{*}{Camera View} & \multicolumn{6}{c}{Tasks (\textbf{New Environment})}  \\
& Bottle placing & Rope boxing  & Shirt folding & Block disassembly & Take Drink from Bag& Average \\
\midrule
UMI &30\%  &0\% &0\% &0\% &0\% &6\% \\
UMI w/ Fixed Head Camera &30\% &10\% &20\% &0\% &20\% & 16\%\\
\textbf{ActiveUMI}      & \textbf{70\%}  & \textbf{50\%}  & \textbf{80\%}  & \textbf{30\%}   & \textbf{50\%} & \textbf{56\%} \\
\bottomrule
\end{tabular}}
\end{table*}

\subsection{Implementation Details and Task Descriptions} 
Our real-world experiments are conducted on a testbed consisting of three 6-DoF ARX R5 robotic arms.  Two arms, each equipped with a fisheye wrist-mounted camera, form a bimanual manipulation system. The third arm provides an active, mobile viewpoint, with its camera feed sourced from a human operator's VR headset to simulate an egocentric head camera. All sensor and robot data is collected at a frequency of 30Hz. For policy learning, we uses $\pi_{0}$, a state-of-the-art vision-language-action (VLA) model. For the fine-tuning stage, the model is subsequently fine-tuned for 50k iterations using a cosine learning rate scheduler. Unless otherwise stated, all experiments were conducted over 10 trials.

Our approach was evaluated on a diverse set of tasks, each designed to test a different robotic skill set:
\begin{itemize}
    \item \textbf{Block disassembly}: A precision task requiring the robot to separate two small, interlocked blocks and sort them into a box.
    \item \textbf{Shirt folding}: A deformable object manipulation task demanding accurate state recognition to correctly fold the cloth.
    \item \textbf{Rope boxing}: A long-horizon task where the robot must neatly guide a long rope into a box.
    \item \textbf{Toolbox cleaning}: An articulated object manipulation task requiring the robot to operate a hinge to close the lid of a toolbox.
    \item \textbf{Bottle placing}: A task designed to test the policy's generalization and robustness to significant randomization in object positions.
\end{itemize}
We give an example for each task in Figure~\ref{fig:tasks}.

\subsection{How Important is the Egocentric Active Perception?}
A key feature of our proposed ActiveUMI framework is its use of active perception. This section investigates the impact of this component on model performance for complex manipulation tasks. Specifically, we compare the following three experimental setups:
\begin{itemize}
    \item \textbf{Active Perception} (Our Method): The full ActiveUMI system, which includes a mobile head camera controlled by a dedicated 6-DoF arm (total 20-DoF).
    \item \textbf{Fixed Head Camera}: A baseline where the head camera is mounted in a static, top-down position, removing the active perception component (total 14-DoF).
    \item \textbf{Wrist-Camera-Only} (UMI Baseline): A second baseline where the head camera is removed entirely, leaving only the two fisheye wrist cameras. This configuration replicates the standard setup of UMI-style methods (total 14-DoF).
\end{itemize}

We use $\pi_{0}$ as the base model to train policies for all three configurations. For the wrist-camera-only baseline, we follow the official pi0 implementation and pad the visual tokens corresponding to the missing third-camera view. The experimental results, demonstrated in Table~\ref{tbl:main_indomain}, show that equipping the agent with active perception significantly outperforms both counterparts on all evaluated tasks. For instance, on the PourWater task, our method achieves a success rate 30\% higher than the fixed top-down camera setup and 60\% higher than the wrist-camera-only baseline.

We hypothesize two drivers of the improvements: (i) during in‑the‑wild data collection, demonstrators move their head and body; an active camera lets the policy compensate for this motion rather than treat it as observation noise; and (ii) active viewpoint selection enables the policy to acquire task‑critical information (e.g., verifying a grasp) on demand. Finally, the fixed top‑down camera reliably outperforms wrist‑only, indicating that a third‑person view adds complementary information for complex bimanual tasks.

\begin{figure*}[h!t]
\centerline{\includegraphics[width=1\linewidth]{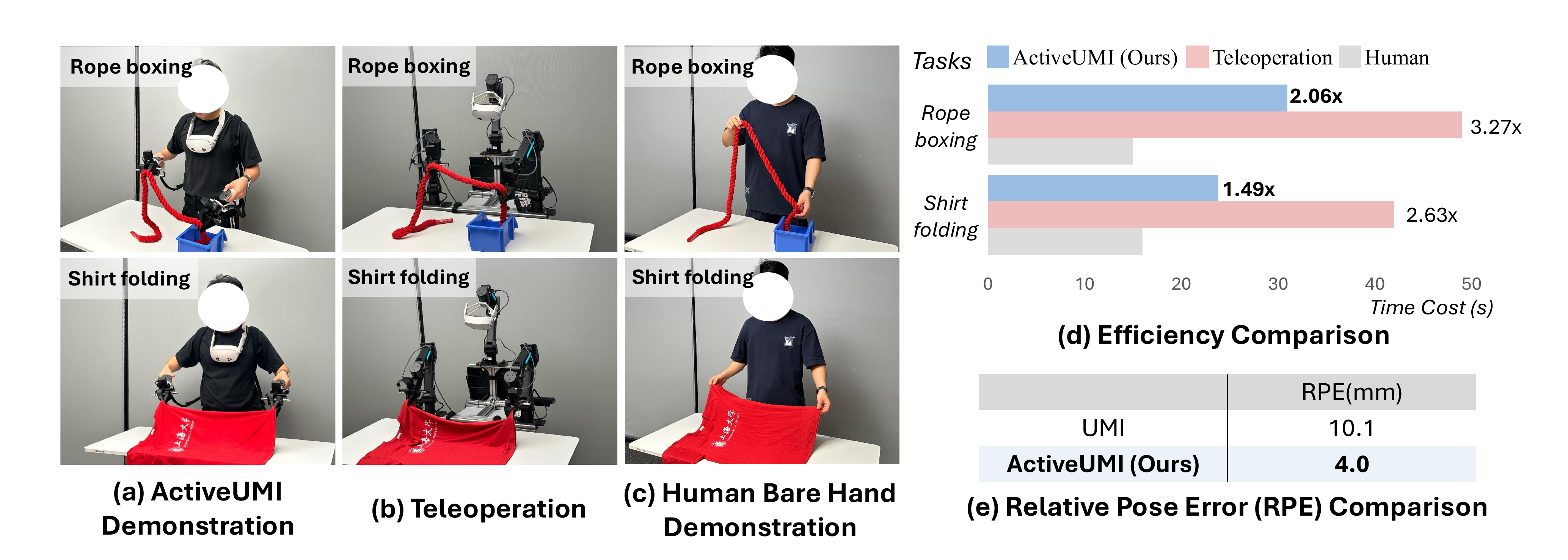}}
\caption{\textbf{Data Collection Comparison.} (a)-(d) We utilize efficiency comparison among ActivateUMI, bare hand, and teleoperation in two tasks: rope boxing and shirt folding. ActivateUMI reaches an efficiency level between bare hand and teleoperation, and consistently outperforms teleoperation across both tasks.} (e) The comparison of relative pose error between UMI and ActiveUMI. 
\label{tbl:efficiency_and_rpe}
\vspace{-1em}
\end{figure*}

\subsection{Mixed Training with Teleoperated Data}
This section investigates the optimal strategy for using ActiveUMI data in policy training. To address the visual and embodiment gaps between human demonstrations and the robot, we evaluated a mixed-data approach on the complex, long-horizon shirt-folding task, conducting 20 trials for each experiment. Specifically, we compared three configurations: (1) training exclusively on ActiveUMI data, (2) mixing ActiveUMI data with 10\% teleoperated data, and (3) mixing ActiveUMI data with only 1\% teleoperated data.

The results, shown in Table~\ref{tbl:data_mixing}, indicate that adding teleoperated data improves performance. For instance, adding 10\% teleoperated data increased the success rate from 80\% to 90\%. Interestingly, the optimal strategy was mixing in just 1\% teleoperated data, which achieved a 95\% success rate. This finding aligns with previous work showing that policies can be trained effectively by combining large-scale simulated data with a small amount of real-world demonstrations. This suggests that we can leverage large-scale, low-cost ActiveUMI data for effective model training, significantly lowering the cost of developing robot foundation models.

This demonstrates that ActiveUMI data is highly sample-efficient, requiring only a small fraction of real-world data to significantly boost and fine-tune the policy's performance. This conclusion aligns with findings from previous work, which have shown that policies can be effectively trained by mixing large-scale data with very few real-world teleoperated demonstrations.

\begin{table}[t]
\centering
\caption{\textbf{Data Mixing Ratio Experiments.} We conducted experiments on the shirt folding task to find the optimal data mixture for maximizing model performance.}
\label{tbl:data_mixing}
\resizebox{0.35\textwidth}{!}{\begin{tabular}{c|c|c|c}
\toprule
Teleoperated Data Ratio & 10\% & 1\%  &  0\% \\
\midrule
Avg. Success Rate & 90\% & 95\% & 80\% \\
\bottomrule

\end{tabular}}
\vspace{-1em}
\end{table}

\subsection{Generalization Capability of ActiveUMI for In-the-Wild Data Collection}
A key indicator of a robust policy is its ability to generalize to novel objects and unseen scenes. To evaluate this capability, we tested the policies trained on ActiveUMI data in a new environment, performing the same set of tasks as the in-domain evaluation. 

This experiment aims to determine if the skills learned, particularly active perception, can transfer to a different visual context. The results, presented in Table III, show that the policy trained with ActiveUMI demonstrates strong generalization capabilities. It achieves an average success rate of 56\% in the new environment, retaining a significant portion of its in-domain performance.

Crucially, this performance significantly surpasses the baselines in the novel setting. The policy using a fixed head camera dropped to a 16\% success rate, while the wrist-camera-only (UMI) baseline's performance fell to just 6\%. This indicates that policies relying on more static or constrained viewpoints fail to adapt when the environment changes. In contrast, the ability to actively control its viewpoint allows the ActiveUMI policy to be more resilient to visual shifts. These findings validate that the ``in-the-wild" data from ActiveUMI, enriched with active perception, produces policies that are not only effective but also generalizable.

\subsection{Data Collection Throughput and Accuracy}
\label{sec:throughput_and_accuracy}
\textbf{Throughput.} The previous section demonstrated that data collected by ActiveUMI is effective for training policies with active perception. A key advantage of our approach is its data collection efficiency. To evaluate this, we measured the time required to complete two long-horizon tasks—rope boxing and shirt folding—using three distinct methods: ActiveUMI, teleoperation of a real robot via a VR kit, and direct human demonstration.

As shown in Figure~\ref{tbl:efficiency_and_rpe}(d), ActiveUMI significantly speeds up data collection compared to teleoperation. For the rope boxing task, ActiveUMI was 2.06x slower than a direct human demonstration, while conventional teleoperation was 3.27x slower. Similarly, for shirt folding, ActiveUMI was 1.49x slower, compared to 2.63x for teleoperation. These results highlight that ActiveUMI provides a practical middle ground, retaining much of the efficiency of natural human motion while being substantially faster than conventional teleoperation.

\textbf{Data Collection Accuracy.} Having shown that ActiveUMI is effective as both a sole training data source and a supplement to teleoperated data, this section evaluates its collection accuracy. Specifically, we measure the error between the data collected by ActiveUMI and the actual trajectories replayed by the robot.

Specifically, we measure the Relative Pose Error (RPE). The experimental task was as follows: an operator, holding the ActiveUMI controller's gripper, placed the gripper at both ends of the tape measure, recording the nominal distance manually. The nominal distances started from \SI{100}{cm} and decreased in \SI{10}{cm} steps, sequentially collecting data for \SI{100}{cm}, \SI{90}{cm}, \ldots, \SI{10}{cm}, for a total of 10 data points. During the experiment, the 6DoF pose sequences of the two grippers were recorded in real-time. We then analyzed the positioning accuracy of the ActiveUMI system based on this recorded data. Next, we entered the playback phase, where the saved pose sequences were precisely replicated on a real robot. At this point, we used the same tape measure to measure the actual distance between the inside of the two grippers, which was recorded as the \textit{playback distance}. Using the nominal distance as the ground truth, we calculated the absolute error of the playback distance relative to the nominal distance:

\begin{equation}
     \Delta L = \left| L_{\text{replay}} - L_{\text{measure}} \right|.
\end{equation}
We further computed the relative error as:
\begin{equation}
    RPE = \frac{\Delta L}{L_{\text{measure}}} \times 100\%.
\end{equation}

We record the average RPE of ten trials and compares with the UMI. The experimental results are shown in Figure~\ref{tbl:efficiency_and_rpe}(e). We can observe that the RPE of UMI is 2.5x smaller than UMI. This low error is naturally comes from the advantange of the VR system, thus we obtain much better data quality and train train good policy network.

\section{CONCLUSIONS}
In conclusion, we identified a critical limitation in current robot data collection methods: the neglect of active, egocentric perception. While humans naturally move their heads to understand and interact with the world, most robot learning systems rely on action-centric, wrist-mounted cameras that limit performance on complex, long-horizon, or occluded tasks. To address this, we introduced ActiveUMI, a portable, in-the-wild data collection framework that couples high-fidelity embodiment alignment with learned viewpoint control. Our experiments demonstrate that this approach is highly effective. Policies trained exclusively on ActiveUMI data achieve a 70\% success rate on a variety of challenging bimanual tasks. Crucially, our method significantly outperforms baselines that lack active perception, confirming that learning how to look is as important as learning what to do. The strong generalization performance on novel objects and scenes further validates the quality of in-the-wild data collected with this approach.

{
    \small
    \bibliographystyle{ieeenat_fullname}
    \bibliography{main}
}


\end{document}


%% file: main.bbl
\begin{thebibliography}{54}
\providecommand{\natexlab}[1]{#1}
\providecommand{\url}[1]{\texttt{#1}}
\expandafter\ifx\csname urlstyle\endcsname\relax
  \providecommand{\doi}[1]{doi: #1}\else
  \providecommand{\doi}{doi: \begingroup \urlstyle{rm}\Url}\fi

\bibitem[Arachchige et~al.(2025)Arachchige, Chen, Jung, Shin, Bansal, Barroso, He, Lin, Joffe, Kousik, et~al.]{arachchige2025sail}
Nadun~Ranawaka Arachchige, Zhenyang Chen, Wonsuhk Jung, Woo~Chul Shin, Rohan Bansal, Pierre Barroso, Yu~Hang He, Yingyang~Celine Lin, Benjamin Joffe, Shreyas Kousik, et~al.
\newblock Sail: Faster-than-demonstration execution of imitation learning policies.
\newblock \emph{arXiv preprint arXiv:2506.11948}, 2025.

\bibitem[Bahl et~al.(2022)Bahl, Gupta, and Pathak]{bahl2022human}
Shikhar Bahl, Abhinav Gupta, and Deepak Pathak.
\newblock Human-to-robot imitation in the wild.
\newblock \emph{arXiv preprint arXiv:2207.09450}, 2022.

\bibitem[Ben et~al.(2025)Ben, Jia, Zeng, Dong, Lin, and Pang]{ben2024homie}
Qingwei Ben, Feiyu Jia, Jia Zeng, Junting Dong, Dahua Lin, and Jiangmiao Pang.
\newblock Homie: Humanoid loco-manipulation with isomorphic exoskeleton cockpit.
\newblock \emph{arXiv preprint arXiv:2502.13013}, 2025.

\bibitem[Black et~al.(2024)Black, Brown, Driess, et~al.]{black2024pi0}
Kevin Black, Noah Brown, Danny Driess, et~al.
\newblock $\pi_0$: A vision-language-action flow model for general robot control.
\newblock \emph{arXiv preprint arXiv:2410.24164}, 2024.

\bibitem[Chen et~al.(2025{\natexlab{a}})Chen, Liu, Gu, Liu, Zhang, Li, He, Guo, Fu, Zhang, et~al.]{chen2025fast}
Hao Chen, Jiaming Liu, Chenyang Gu, Zhuoyang Liu, Renrui Zhang, Xiaoqi Li, Xiao He, Yandong Guo, Chi-Wing Fu, Shanghang Zhang, et~al.
\newblock Fast-in-slow: A dual-system foundation model unifying fast manipulation within slow reasoning.
\newblock \emph{arXiv preprint arXiv:2506.01953}, 2025{\natexlab{a}}.

\bibitem[Chen et~al.(2025{\natexlab{b}})Chen, Liu, Qian, Jiang, Li, Zhang, Liu, Gu, Hou, Wang, et~al.]{chen2025acdit}
Sixiang Chen, Jiaming Liu, Siyuan Qian, Han Jiang, Lily Li, Renrui Zhang, Zhuoyang Liu, Chenyang Gu, Chengkai Hou, Pengwei Wang, et~al.
\newblock Ac-dit: Adaptive coordination diffusion transformer for mobile manipulation.
\newblock \emph{arXiv preprint arXiv:2507.01961}, 2025{\natexlab{b}}.

\bibitem[Cheng et~al.(2024)Cheng, Li, Yang, Yang, and Wang]{cheng2024opentelevision}
Xuxin Cheng, Jialong Li, Shiqi Yang, Ge Yang, and Xiaolong Wang.
\newblock Open-television: Teleoperation with immersive active visual feedback.
\newblock \emph{arXiv preprint arXiv:2407.01512}, 2024.

\bibitem[Chi et~al.(2023)Chi, Feng, Du, Xu, Cousineau, Burchfiel, and Song]{chi2023diffusion_policy}
Cheng Chi, Siyuan Feng, Yilun Du, Zhenjia Xu, Eric Cousineau, Benjamin Burchfiel, and Shuran Song.
\newblock Diffusion policy: Visuomotor policy learning via action diffusion.
\newblock \emph{RSS}, 2023.

\bibitem[Chi et~al.(2024)Chi, Xu, Pan, Cousineau, Burchfiel, Feng, Tedrake, and Song]{chi2024umi}
Cheng Chi, Zhenjia Xu, Chuer Pan, Eric Cousineau, Benjamin Burchfiel, Siyuan Feng, Russ Tedrake, and Shuran Song.
\newblock Universal manipulation interface: In-the-wild robot teaching without in-the-wild robots.
\newblock \emph{arXiv preprint arXiv:2402.10329}, 2024.

\bibitem[Dasari et~al.(2024)Dasari, Mees, Zhao, Srirama, and Levine]{dasari2024ditpi}
Sudeep Dasari, Oier Mees, Sebastian Zhao, Mohan~Kumar Srirama, and Sergey Levine.
\newblock The ingredients for robotic diffusion transformers.
\newblock \emph{arXiv preprint arXiv:2410.10088}, 2024.

\bibitem[Deng et~al.(2025)Deng, Yan, Wei, Ma, Yang, Chen, Zhang, Yang, Zhang, Zhang, et~al.]{deng2025graspvla}
Shengliang Deng, Mi Yan, Songlin Wei, Haixin Ma, Yuxin Yang, Jiayi Chen, Zhiqi Zhang, Taoyu Yang, Xuheng Zhang, Wenhao Zhang, et~al.
\newblock Graspvla: a grasping foundation model pre-trained on billion-scale synthetic action data.
\newblock \emph{arXiv preprint arXiv:2505.03233}, 2025.

\bibitem[Ding et~al.(2025)Ding, Ma, Tong, Zou, Luo, Fan, Wang, Lu, Mo, Liu, et~al.]{ding2025humanoidvla}
Pengxiang Ding, Jianfei Ma, Xinyang Tong, Binghong Zou, Xinxin Luo, Yiguo Fan, Ting Wang, Hongchao Lu, Panzhong Mo, Jinxin Liu, et~al.
\newblock Humanoid-vla: Towards universal humanoid control with visual integration.
\newblock \emph{arXiv preprint arXiv:2502.14795}, 2025.

\bibitem[Fang et~al.(2024)Fang, Fang, Wang, Ren, Chen, Zhang, Wang, and Lu]{fang2024airexo}
Hongjie Fang, Hao-Shu Fang, Yiming Wang, Jieji Ren, Jingjing Chen, Ruo Zhang, Weiming Wang, and Cewu Lu.
\newblock Airexo: Low-cost exoskeletons for learning whole-arm manipulation in the wild.
\newblock \emph{arXiv preprint arXiv:2309.14975}, 2024.

\bibitem[Fang et~al.(2025{\natexlab{a}})Fang, Wang, Wang, Chen, Xia, Lv, He, Yi, Guo, Zhan, Yang, Wang, Lu, and Fang]{fang2025airexo2}
Hongjie Fang, Chenxi Wang, Yiming Wang, Jingjing Chen, Shangning Xia, Jun Lv, Zihao He, Xiyan Yi, Yunhan Guo, Xinyu Zhan, Lixin Yang, Weiming Wang, Cewu Lu, and Hao-Shu Fang.
\newblock Airexo-2: Scaling up generalizable robotic imitation learning with low-cost exoskeletons.
\newblock \emph{arXiv preprint arXiv:2503.03081}, 2025{\natexlab{a}}.

\bibitem[Fang et~al.(2025{\natexlab{b}})Fang, Romero, Xie, Hu, Huang, Alvarez, Kim, Margolis, Anbarasu, Tomizuka, et~al.]{fang2025dexop}
Hao-Shu Fang, Branden Romero, Yichen Xie, Arthur Hu, Bo-Ruei Huang, Juan Alvarez, Matthew Kim, Gabriel Margolis, Kavya Anbarasu, Masayoshi Tomizuka, et~al.
\newblock Dexop: A device for robotic transfer of dexterous human manipulation.
\newblock \emph{arXiv preprint arXiv:2509.04441}, 2025{\natexlab{b}}.

\bibitem[Fu et~al.(2024)Fu, Zhao, Wu, Wetzstein, and Finn]{fu2024humanplus}
Zipeng Fu, Qingqing Zhao, Qi Wu, Gordon Wetzstein, and Chelsea Finn.
\newblock Humanplus: Humanoid shadowing and imitation from humans.
\newblock \emph{arXiv preprint arXiv:2406.10454}, 2024.

\bibitem[Geng et~al.(2025)Geng, Wang, Wei, Li, Wang, An, Cheng, Lou, Li, Wang, et~al.]{geng2025roboverse}
Haoran Geng, Feishi Wang, Songlin Wei, Yuyang Li, Bangjun Wang, Boshi An, Charlie~Tianyue Cheng, Haozhe Lou, Peihao Li, Yen-Jen Wang, et~al.
\newblock Roboverse: Towards a unified platform, dataset and benchmark for scalable and generalizable robot learning.
\newblock \emph{arXiv preprint arXiv:2504.18904}, 2025.

\bibitem[Grauman et~al.(2022)Grauman, Westbury, Byrne, Chavis, Furnari, Girdhar, Hamburger, Jiang, Liu, Liu, et~al.]{grauman2022ego4d}
Kristen Grauman, Andrew Westbury, Eugene Byrne, Zachary Chavis, Antonino Furnari, Rohit Girdhar, Jackson Hamburger, Hao Jiang, Miao Liu, Xingyu Liu, et~al.
\newblock Ego4d: Around the world in 3,000 hours of egocentric video.
\newblock In \emph{Proceedings of the IEEE/CVF Conference on Computer Vision and Pattern Recognition}, pages 18995--19012, 2022.

\bibitem[He et~al.(2024)He, Luo, He, Xiao, Zhang, Zhang, Kitani, Liu, and Shi]{he2024omnih2o}
Tairan He, Zhengyi Luo, Xialin He, Wenli Xiao, Chong Zhang, Weinan Zhang, Kris Kitani, Changliu Liu, and Guanya Shi.
\newblock Omnih2o: Universal and dexterous human-to-humanoid whole-body teleoperation and learning.
\newblock \emph{arXiv preprint arXiv:2406.08858}, 2024.

\bibitem[Hu et~al.(2024)Hu, Guo, Wang, Chen, Wang, Zhang, Sreenath, Lu, and Chen]{hu2024vpp}
Yucheng Hu, Yanjiang Guo, Pengchao Wang, Xiaoyu Chen, Yen-Jen Wang, Jianke Zhang, Koushil Sreenath, Chaochao Lu, and Jianyu Chen.
\newblock Video prediction policy: A generalist robot policy with predictive visual representations.
\newblock \emph{arXiv preprint arXiv:2412.14803}, 2024.

\bibitem[Jia et~al.(2024)Jia, Liu, Chen, Gu, Wang, Luo, Lee, Wang, Wang, Zhang, et~al.]{jia2024lift3d}
Yueru Jia, Jiaming Liu, Sixiang Chen, Chenyang Gu, Zhilue Wang, Longzan Luo, Lily Lee, Pengwei Wang, Zhongyuan Wang, Renrui Zhang, et~al.
\newblock Lift3d foundation policy: Lifting 2d large-scale pretrained models for robust 3d robotic manipulation.
\newblock \emph{arXiv preprint arXiv:2411.18623}, 2024.

\bibitem[Jiang et~al.(2025)Jiang, Xie, Lin, Xu, Wan, Mandlekar, Fan, and Zhu]{jiang2025dexmimicgen}
Zhenyu Jiang, Yuqi Xie, Kevin Lin, Zhenjia Xu, Weikang Wan, Ajay Mandlekar, Linxi~Jim Fan, and Yuke Zhu.
\newblock Dexmimicgen: Automated data generation for bimanual dexterous manipulation via imitation learning.
\newblock In \emph{2025 IEEE International Conference on Robotics and Automation (ICRA)}, pages 16923--16930. IEEE, 2025.

\bibitem[Kareer et~al.(2025)Kareer, Patel, Punamiya, Mathur, Cheng, Wang, Hoffman, and Xu]{Kareer2025EgoMimic}
Simar Kareer, Dhruv Patel, Ryan Punamiya, Pranay Mathur, Shuo Cheng, Chen Wang, Judy Hoffman, and Danfei Xu.
\newblock Egomimic: Scaling imitation learning via egocentric video.
\newblock In \emph{2025 IEEE International Conference on Robotics and Automation (ICRA)}, pages 13226--13233, 2025.

\bibitem[Li et~al.(2025{\natexlab{a}})Li, Wen, Peng, Peng, Feng, and Zhu]{li2025pointvla}
Chengmeng Li, Junjie Wen, Yan Peng, Yaxin Peng, Feifei Feng, and Yichen Zhu.
\newblock Pointvla: Injecting the 3d world into vision-language-action models.
\newblock \emph{arXiv preprint arXiv:2503.07511}, 2025{\natexlab{a}}.

\bibitem[Li et~al.(2025{\natexlab{b}})Li, Wu, Xi, Li, Huang, Zhang, Chen, Wang, Zhu, Liu, et~al.]{li2025controlvla}
Puhao Li, Yingying Wu, Ziheng Xi, Wanlin Li, Yuzhe Huang, Zhiyuan Zhang, Yinghan Chen, Jianan Wang, Song-Chun Zhu, Tengyu Liu, et~al.
\newblock Controlvla: Few-shot object-centric adaptation for pre-trained vision-language-action models.
\newblock \emph{arXiv preprint arXiv:2506.16211}, 2025{\natexlab{b}}.

\bibitem[Li et~al.(2025{\natexlab{c}})Li, Lin, Cui, Liu, Liang, Zhu, and Huang]{li2025clone}
Yixuan Li, Yutang Lin, Jieming Cui, Tengyu Liu, Wei Liang, Yixin Zhu, and Siyuan Huang.
\newblock Clone: Closed-loop whole-body humanoid teleoperation for long-horizon tasks.
\newblock \emph{arXiv preprint arXiv:2506.08931}, 2025{\natexlab{c}}.

\bibitem[Liu et~al.(2025{\natexlab{a}})Liu, Chen, An, Liu, Zhang, Gu, Li, Guo, Chen, Liu, et~al.]{liu2025hybridvla}
Jiaming Liu, Hao Chen, Pengju An, Zhuoyang Liu, Renrui Zhang, Chenyang Gu, Xiaoqi Li, Ziyu Guo, Sixiang Chen, Mengzhen Liu, et~al.
\newblock Hybridvla: Collaborative diffusion and autoregression in a unified vision-language-action model.
\newblock \emph{arXiv preprint arXiv:2503.10631}, 2025{\natexlab{a}}.

\bibitem[Liu et~al.(2024)Liu, Guan, Jia, Wu, Liu, Wang, Liang, Chen, Zhang, Song, et~al.]{liu2024fastumi}
Kehui Liu, Chuyue Guan, Zhongjie Jia, Ziniu Wu, Xin Liu, Tianyu Wang, Shuai Liang, Pengan Chen, Pingrui Zhang, Haoming Song, et~al.
\newblock Fastumi: A scalable and hardware-independent universal manipulation interface with dataset.
\newblock \emph{arXiv preprint arXiv:2409.19499}, 2024.

\bibitem[Liu et~al.(2025{\natexlab{b}})Liu, Shin, Han, Chen, Ravichandar, and Xu]{liu2025immimic}
Yangcen Liu, Woo~Chul Shin, Yunhai Han, Zhenyang Chen, Harish Ravichandar, and Danfei Xu.
\newblock Immimic: Cross-domain imitation from human videos via mapping and interpolation.
\newblock \emph{arXiv preprint arXiv:2509.10952}, 2025{\natexlab{b}}.

\bibitem[Maddukuri et~al.(2025)Maddukuri, Jiang, Chen, Nasiriany, Xie, Fang, Huang, Wang, Xu, Chernyadev, et~al.]{maddukuri2025simandreal}
Abhiram Maddukuri, Zhenyu Jiang, Lawrence~Yunliang Chen, Soroush Nasiriany, Yuqi Xie, Yu Fang, Wenqi Huang, Zu Wang, Zhenjia Xu, Nikita Chernyadev, et~al.
\newblock Sim-and-real co-training: A simple recipe for vision-based robotic manipulation.
\newblock \emph{arXiv preprint arXiv:2503.24361}, 2025.

\bibitem[NVIDIA(2025)]{nvidia2025gr00tn1}
NVIDIA.
\newblock Gr00t n1: An open foundation model for generalist humanoid robots.
\newblock \emph{arXiv preprint arXiv:2503.14734}, 2025.

\bibitem[{Physical Intelligence}(2025)]{intelligence2025pi05}
{Physical Intelligence}.
\newblock $\pi_{0.5}$: a vision-language-action model with open-world generalization.
\newblock \emph{arXiv preprint arXiv:2504.16054}, 2025.

\bibitem[Punamiya et~al.()Punamiya, Patel, Aphiwetsa, Kuppili, Zhu, Kareer, Hoffman, and Xu]{punamiya2025egobridge}
Ryan Punamiya, Dhruv Patel, Patcharapong Aphiwetsa, Pranav Kuppili, Lawrence~Y Zhu, Simar Kareer, Judy Hoffman, and Danfei Xu.
\newblock Egobridge: Domain adaptation for generalizable imitation from egocentric human data.
\newblock In \emph{Human to Robot: Workshop on Sensorizing, Modeling, and Learning from Humans}.

\bibitem[Qiu et~al.(2025)Qiu, Yang, Cheng, Chawla, Li, He, Yan, Yoon, Hoque, Paulsen, Yang, Zhang, Yi, Shi, and Wang]{qiu2025humanoidpolicy}
Ri-Zhao Qiu, Shiqi Yang, Xuxin Cheng, Chaitanya Chawla, Jialong Li, Tairan He, Ge Yan, David~J. Yoon, Ryan Hoque, Lars Paulsen, Ge Yang, Jian Zhang, Sha Yi, Guanya Shi, and Xiaolong Wang.
\newblock Humanoid policy - human policy.
\newblock \emph{arXiv preprint arXiv:2503.13441}, 2025.

\bibitem[Reuss et~al.()Reuss, Ya{\u{g}}murlu, Wenzel, and Lioutikov]{reuss2024multimodal}
Moritz Reuss, {\"O}mer~Erdin{\c{c}} Ya{\u{g}}murlu, Fabian Wenzel, and Rudolf Lioutikov.
\newblock Multimodal diffusion transformer: Learning versatile behavior from multimodal goals.
\newblock In \emph{First Workshop on Vision-Language Models for Navigation and Manipulation at ICRA 2024}.

\bibitem[Srirama et~al.(2024)Srirama, Dasari, Bahl, and Gupta]{kumar2024hrp}
Mohan~Kumar Srirama, Sudeep Dasari, Shikhar Bahl, and Abhinav Gupta.
\newblock Hrp: Human affordances for robotic pre-training.
\newblock In \emph{Proceedings of Robotics: Science and Systems}, Delft, Netherlands, 2024.

\bibitem[Sun et~al.(2025)Sun, Feng, Cao, Liu, Jin, and Xie]{sun2025ulc}
Wandong Sun, Luying Feng, Baoshi Cao, Yang Liu, Yaochu Jin, and Zongwu Xie.
\newblock Ulc: A unified and fine-grained controller for humanoid loco-manipulation.
\newblock \emph{arXiv preprint arXiv:2507.06905}, 2025.

\bibitem[Team et~al.(2025)Team, Abeyruwan, Ainslie, Alayrac, Arenas, Armstrong, Balakrishna, Baruch, Bauza, Blokzijl, et~al.]{team2025gemini}
Gemini~Robotics Team, Saminda Abeyruwan, Joshua Ainslie, Jean-Baptiste Alayrac, Montserrat~Gonzalez Arenas, Travis Armstrong, Ashwin Balakrishna, Robert Baruch, Maria Bauza, Michiel Blokzijl, et~al.
\newblock Gemini robotics: Bringing ai into the physical world.
\newblock \emph{arXiv preprint arXiv:2503.20020}, 2025.

\bibitem[Wang et~al.(2024)Wang, Shi, Wang, Zhang, Fei-Fei, and Liu]{wang2024dexcap}
Chen Wang, Haochen Shi, Weizhuo Wang, Ruohan Zhang, Li Fei-Fei, and C.~Karen Liu.
\newblock Dexcap: Scalable and portable mocap data collection system for dexterous manipulation.
\newblock \emph{arXiv preprint arXiv:2403.07788}, 2024.

\bibitem[Wang et~al.(2025)Wang, Zhang, Li, Liu, Li, Wu, Zhong, Yu, Zhang, and Wang]{wang2025trackvla}
Shaoan Wang, Jiazhao Zhang, Minghan Li, Jiahang Liu, Anqi Li, Kui Wu, Fangwei Zhong, Junzhi Yu, Zhizheng Zhang, and He Wang.
\newblock Trackvla: Embodied visual tracking in the wild.
\newblock \emph{arXiv preprint arXiv:2505.23189}, 2025.

\bibitem[Wen et~al.(2024)Wen, Zhu, Li, Zhu, Wu, Xu, Cheng, Shen, Peng, Feng, et~al.]{wen2024tinyvla}
Junjie Wen, Yichen Zhu, Jinming Li, Minjie Zhu, Kun Wu, Zhiyuan Xu, Ran Cheng, Chaomin Shen, Yaxin Peng, Feifei Feng, et~al.
\newblock Tinyvla: Towards fast, data-efficient vision-language-action models for robotic manipulation.
\newblock \emph{arXiv preprint arXiv:2409.12514}, 2024.

\bibitem[Wen et~al.(2025)Wen, Zhu, Li, Tang, Shen, and Feng]{wen2025dexvla}
Junjie Wen, Yichen Zhu, Jinming Li, Zhibin Tang, Chaomin Shen, and Feifei Feng.
\newblock Dexvla: Vision-language model with plug-in diffusion expert for general robot control.
\newblock \emph{arXiv preprint arXiv:2502.05855}, 2025.

\bibitem[Xiong et~al.(2025)Xiong, Xu, Wu, Hou, Bohg, and Song]{xiong2025vision}
Haoyu Xiong, Xiaomeng Xu, Jimmy Wu, Yifan Hou, Jeannette Bohg, and Shuran Song.
\newblock Vision in action: Learning active perception from human demonstrations.
\newblock \emph{arXiv preprint arXiv:2506.15666}, 2025.

\bibitem[Xu et~al.(2025)Xu, Zhang, Hou, Xu, Fan, Veloso, and Song]{xu2025dexumi}
Mengda Xu, Han Zhang, Yifan Hou, Zhenjia Xu, Linxi Fan, Manuela Veloso, and Shuran Song.
\newblock Dexumi: Using human hand as the universal manipulation interface for dexterous manipulation.
\newblock \emph{arXiv preprint arXiv:2505.21864}, 2025.

\bibitem[Yang et~al.(2025)Yang, Yu, Wu, Yan, Li, Cheng, Zou, Fang, Cheng, Qiu, et~al.]{yang2025egovla}
Ruihan Yang, Qinxi Yu, Yecheng Wu, Rui Yan, Borui Li, An-Chieh Cheng, Xueyan Zou, Yunhao Fang, Xuxin Cheng, Ri-Zhao Qiu, et~al.
\newblock Egovla: Learning vision-language-action models from egocentric human videos.
\newblock \emph{arXiv preprint arXiv:2507.12440}, 2025.

\bibitem[Ye et~al.(2025)Ye, Liu, Ding, Gao, Rybkin, and Abbeel]{ye2025video2policy}
Weirui Ye, Fangchen Liu, Zheng Ding, Yang Gao, Oleh Rybkin, and Pieter Abbeel.
\newblock Video2policy: Scaling up manipulation tasks in simulation through internet videos.
\newblock \emph{arXiv preprint arXiv:2502.09886}, 2025.

\bibitem[Ze et~al.(2024)Ze, Zhang, Zhang, Hu, Wang, and Xu]{ze20243d}
Yanjie Ze, Gu Zhang, Kangning Zhang, Chenyuan Hu, Muhan Wang, and Huazhe Xu.
\newblock 3d diffusion policy: Generalizable visuomotor policy learning via simple 3d representations.
\newblock In \emph{ICRA 2024 Workshop on 3D Visual Representations for Robot Manipulation}, 2024.

\bibitem[Ze et~al.(2025)Ze, Chen, Araújo, ang Cao, Peng, Wu, and Liu]{ze2025twist}
Yanjie Ze, Zixuan Chen, João~Pedro Araújo, Zi ang Cao, Xue~Bin Peng, Jiajun Wu, and C.~Karen Liu.
\newblock Twist: Teleoperated whole-body imitation system, 2025.

\bibitem[Zhang et~al.(2025)Zhang, Hu, Yuan, and Xu]{zhang2025doglove}
Han Zhang, Songbo Hu, Zhecheng Yuan, and Huazhe Xu.
\newblock Doglove: Dexterous manipulation with a low-cost open-source haptic force feedback glove.
\newblock \emph{arXiv preprint arXiv:2502.07730}, 2025.

\bibitem[Zhao et~al.(2023)Zhao, Kumar, Levine, and Finn]{act}
Tony~Z Zhao, Vikash Kumar, Sergey Levine, and Chelsea Finn.
\newblock Learning fine-grained bimanual manipulation with low-cost hardware.
\newblock \emph{arXiv preprint arXiv:2304.13705}, 2023.

\bibitem[Zhong et~al.(2025)Zhong, Cheng, Xu, Wei, Guo, Zhang, Dai, and Lu]{zhong2025nuexo}
Rui Zhong, Chuang Cheng, Junpeng Xu, Yantong Wei, Ce Guo, Daoxun Zhang, Wei Dai, and Huimin Lu.
\newblock Nuexo: A wearable exoskeleton covering all upper limb rom for outdoor data collection and teleoperation of humanoid robots.
\newblock \emph{arXiv preprint arXiv:2503.10554}, 2025.

\bibitem[Zhou et~al.(2025{\natexlab{a}})Zhou, Zhu, Wen, Shen, and Xu]{zhou2025chatvla2}
Zhongyi Zhou, Yichen Zhu, Junjie Wen, Chaomin Shen, and Yi Xu.
\newblock Vision-language-action model with open-world embodied reasoning from pretrained knowledge.
\newblock \emph{arXiv preprint arXiv:2505.21906}, 2025{\natexlab{a}}.

\bibitem[Zhou et~al.(2025{\natexlab{b}})Zhou, Zhu, Zhu, Wen, Liu, Xu, Meng, Cheng, Peng, Shen, et~al.]{zhou2025chatvla}
Zhongyi Zhou, Yichen Zhu, Minjie Zhu, Junjie Wen, Ning Liu, Zhiyuan Xu, Weibin Meng, Ran Cheng, Yaxin Peng, Chaomin Shen, et~al.
\newblock Chatvla: Unified multimodal understanding and robot control with vision-language-action model.
\newblock \emph{arXiv preprint arXiv:2502.14420}, 2025{\natexlab{b}}.

\bibitem[Zhu et~al.(2025)Zhu, Kuppili, Punamiya, Aphiwetsa, Patel, Kareer, Ha, and Xu]{zhu2025emma}
Lawrence~Y Zhu, Pranav Kuppili, Ryan Punamiya, Patcharapong Aphiwetsa, Dhruv Patel, Simar Kareer, Sehoon Ha, and Danfei Xu.
\newblock Emma: Scaling mobile manipulation via egocentric human data.
\newblock \emph{arXiv preprint arXiv:2509.04443}, 2025.

\end{thebibliography}
